\pdfoutput=1

\documentclass[11pt]{article}

\usepackage[]{emnlp2021}

\usepackage{times}
\usepackage{latexsym}
\usepackage{amsfonts}

\usepackage[T1]{fontenc}

\usepackage[utf8]{inputenc}

\usepackage{microtype}

\usepackage{graphicx} 
\usepackage{booktabs}
\usepackage{url}
\usepackage{multirow}
\usepackage{amsmath}
\usepackage{cleveref}
\crefformat{section}{\S#2#1#3}
\crefformat{subsection}{\S#2#1#3}
\crefformat{subsubsection}{\S#2#1#3}

\newcommand{\situatedqa}[0]{\lbl{Situated{QA}}}

\newcommand{\lbl}[1]{\textsc{#1}} 

\usepackage[textsize=scriptsize]{todonotes}

\title{\situatedqa: Incorporating Extra-Linguistic Contexts into QA}

\author{Michael J.Q. Zhang \and Eunsol Choi \\
        Computer Science Department\\
        The University of Texas at Austin \\
        \texttt{\{mjqzhang, eunsol\}@utexas.edu} \\
        }

\begin{document}
\maketitle
\begin{abstract}
Answers to the same question may change depending on the extra-linguistic contexts (when and where the question was asked).
To study this challenge, we introduce \situatedqa, an open-retrieval QA dataset where systems must produce the correct answer to a question given the temporal or geographical context. To construct \situatedqa, we first identify such questions in existing QA datasets. We find that a significant proportion of information seeking questions have context-dependent answers (e.g., roughly 16.5\% of NQ-Open). For such context-dependent questions, we then crowdsource alternative contexts and their corresponding answers. Our study shows that existing models struggle with producing answers that are frequently updated or from uncommon locations. We further quantify how existing models, which are trained on data collected in the past, fail to generalize to answering questions asked in the present, even when provided with an updated evidence corpus (a roughly 15 point drop in accuracy). Our analysis suggests that open-retrieval QA benchmarks should incorporate extra-linguistic context to stay relevant globally and in the future. Our data, code, and datasheet are available at \url{https://situatedqa.github.io/}.

\end{abstract}

\begin{table*}
\footnotesize
\begin{center}
\begin{tabular}{lrrr}
\toprule
Question $q$ & Context Type $c_t$& Context Value $c_v$ & Answer $a$ \\ \midrule
Who composed the music for the first Harry Potter film? &- & - &-\\
What's the biggest country in Europe excluding Russia? & - & - & -\\\midrule
\multirow{2}{*}{How many seasons are there for American Horror Story?} & \multirow{2}{*}{$\textsc{temp}$} & Sep 18, 2019 & 10 \\
& & Sep 13, 2017 & 9 \\ \midrule
\multirow{2}{*}{Who made the most three point shots in the NBA?} & \multirow{2}{*}{$\textsc{temp}$} & 2014 & Ray Allen \\
& & 2005 & Reggie Miller \\ \midrule
\multirow{2}{*}{When was the last time states were created?} & \multirow{2}{*}{$\textsc{geo}$} & Nigeria & 1 October 1996 \\
& & United States & 1959 \\ \midrule
\multirow{2}{*}{Where do we rank among the world's largest cities?} & \multirow{2}{*}{$\textsc{geo}$} & Tokyo & 1st \\
& & Shanghai & 3rd \\
\bottomrule
\end{tabular}
\end{center}
\vspace{-0.8em}
\caption{Examples of how questions interact with geographical and temporal context in \situatedqa. The first two questions are not identified as geographically nor temporally dependent.}
\label{tab:context_examples}
\end{table*}

\begin{figure}[t]
\centering
\includegraphics{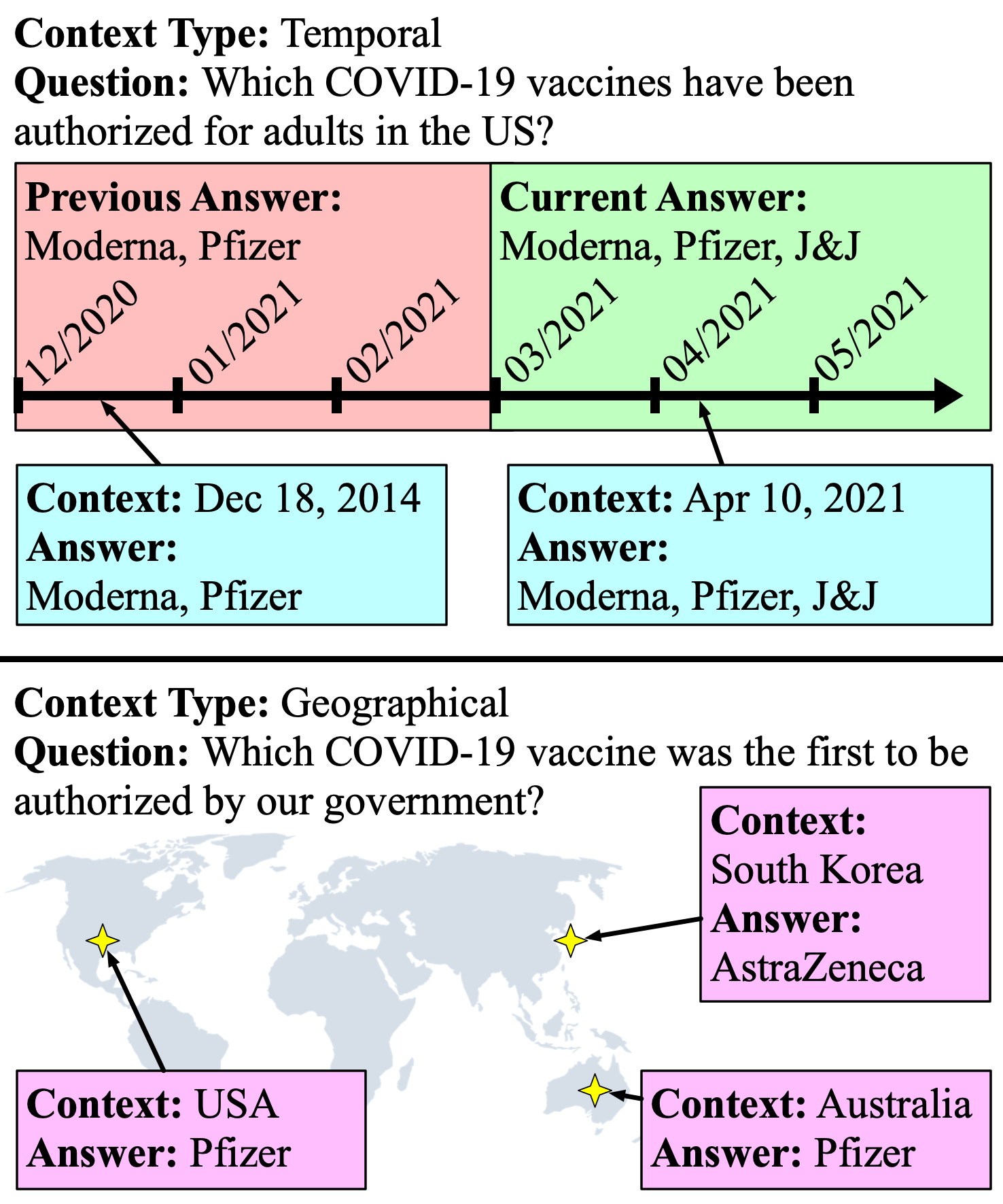}
\vspace{-0.4em}
\caption{Examples of questions with answers that change depending on the temporal or geographical context. }
\vspace{-0.8em}
\label{fig:intro_figure}
\end{figure}

\section{Introduction}\label{sec:intro}

Language reflects our ever-changing world; thus, the meaning of a sentence depends on a variety of extra-linguistic contexts, including \textit{when} it was stated, \textit{who} stated it, and \textit{where} it was stated. Figure~\ref{fig:intro_figure} depicts examples of this, where the answer to a question varies based on the temporal and geographical contexts. In this work, we study such questions by exploring the affects of temporal and geographical contexts in open-retrieval QA.

Ideally, QA systems should provide up-to-date, localized answers to questions like those in Figure~\ref{fig:intro_figure} that are situated in the inquirer's temporal and geographical contexts. The existing paradigm for evaluating QA systems, however, makes implicit assumptions about time and location and does not measure how well QA models adapt to new contexts. Recent work studying other NLP tasks have shown that this static paradigm misrepresents how models perform in practical settings~\cite{osborne2014exponential,Wang2008ContinuousTD,huang-paul-2018-examining,Rijhwani2020TemporallyInformedAO}.
To address this challenge, we introduce \situatedqa, a dataset comprised of questions from existing open-retrieval QA datasets~\cite{Kwiatkowski2019NaturalQA} that have been re-annotated for their their context-dependence and answers across different situations (i.e., temporal or geographical contexts).\footnote{In this work, we do not model temporal \textbf{and} geographical contexts jointly for simplicity.}

As we will see in Section~\ref{sec:collect}, a significant proportion of information-seeking questions are sensitive to the two extra-linguistic contexts we study. We annotate 9K questions from four existing datasets~\cite{Kwiatkowski2019NaturalQA,Berant2013SemanticPO,Clark2020TyDiQA,Campos2016MSMA} with their temporal dependence and 2K questions for their geographical dependence. Annotators find that up to 30\% of questions in existing datasets have answers that change over time, and rule-based heuristics using an NER tagger identify that about 5\% of questions specify a location. We collect answers from different temporal and geographic contexts for such context-dependent questions (see Table~\ref{tab:context_examples}). 

Using our collected data, we evaluate whether existing open-retrieval QA systems adapt to new temporal or geographical contexts by providing answers that are up-to-date or from new locations. While it is often assumed that retrieval based systems for QA~\cite{karpukhin2020dense,guu2020realm} can adapt to updated facts during inference by simply updating the retrieval corpus, we find that this is not the case. State-of-the-art retrieval based systems with updated corpora are 15 percentage points less accurate on questions whose answers have been updated versus those that have remained constant since the time when their large-scale training dataset was collected. We also observe that models fail to generalize to answering questions from new locations~\cite{Shankar2017NoCW}, with accuracy dropping by 10 percentage points when asked to provide the answer from a rare location versus a common one.

To support future research developing open-retrieval QA systems which situate questions within the inquirer's extra-linguistic contexts, we propose two tasks for modeling \textit{what} facts change across different extra-linguistic contexts and \textit{how} those facts change. We also provide fine-grained evaluations for measuring how well models adapt to new temporal and geographical contexts. We establish initial performance levels on our tasks by adapting state-of-the-art methods for open-retrieval QA~\cite{karpukhin2020dense,lewis-etal-2020-bart,2020t5cqba}. Finally, we provide rich analysis of existing QA systems, which suggest that benchmark construction should incorporate extra-linguistic contexts to remain relevant globally and in the future.

\section{Definitions \& Tasks}\label{sec:tasks}
\subsection{Defining Extra-Linguistic Contexts}
We begin by defining the scope of contexts studied in this work. For a given question $q$, we say that $a_i$ is its answer when asked in the context $c_i$. Each context consists of a type $c_{t_i}$ and a value $c_{v_i}$. We study two context types: temporal ($\textsc{temp}$) and geographical ($\textsc{geo}$). \textsc{temp} defines each context value as timestamp (e.g. a date or year) where $a_i$ is the answer to $q$ if it was asked at the time of $c_{v_i}$. \textsc{geo} defines each context value as a geopolitical entity where $a_i$ is the answer to the $q$ in the location $c_{v_i}$. See Table~\ref{tab:context_examples} for examples from each context type. We limit our study to valid questions, ignoring presupposition cases~\cite{Kim2021WhichLI} (e.g., asking who is the CEO of Google before Google was founded).

\subsection{Situated Question Answering}
Given a question $q$ and context $c_i$, the task is to produce the corresponding answer $a_i$ for the provided context. This task requires models to situate the question within the provided extra-linguistic context to produce an appropriate answer. Models are evaluated on exact string match with annotated answer $a_i$ and predicted answer $\hat{a}_i$ after minor normalization~\cite{Rajpurkar2016SQuAD10}.

By evaluating on different sets of extra-linguistic contexts of the same question, we can measure whether models are able to generalize to new contexts. For instance, we can evaluate how models perform on commonly versus rarely asked about locations, or on questions whose answers changed recently or long ago. Furthermore, this task allows us to train systems that explicitly model how facts change across different extra-linguistic contexts.

\subsection{Context-Dependent Question Identification}
In context-dependent question identification, we evaluate whether models can determine whether the answer to a given question depends on its extra-linguistic context. More formally, given a question $q$ and context type $c_t$, models must determine whether there exists two distinct contexts values, $(c_{v_i}, c_{v_j})$, with different respective answers, $a_i \neq a_j$. We cast this as binary classification and evaluate models on their classification accuracy, F1, precision, and recall. 

Identifying what facts change depending on the extra-linguistic contexts is nontrivial even for human annotators that often requires extensive background knowledge on a subject. For example, determining whether the capital of Kazakhstan has changed requires specific knowledge of the nation's history. Identifying these questions has many practical applications, such as identifying suitable questions for QA systems that can only provide static answers, an idea we will explore later in Section~\ref{sec:analysis}.

\begin{figure*}
\centering
\includegraphics[width=16cm]{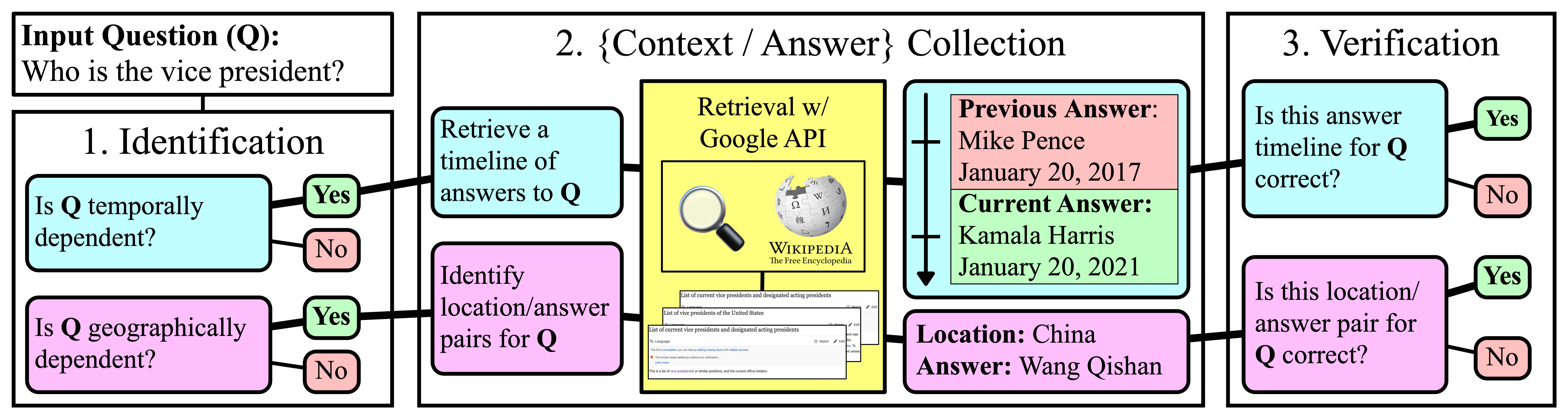}
\caption{Data collection pipeline: Crowdworkers are first asked to identify context dependent questions. We then collect brief answer timelines for temporally dependent questions and location/answer pairs geographically dependent questions, each of which is then verified by another worker.}
\vspace{-0.8em}
\label{fig:data_collection}
\end{figure*}

\section{Data Collection }\label{sec:collect}
Here we describe the process for identifying context-dependent questions and collecting answers from alternate contexts. We split up data collection into three stages which are depicted in Figure~\ref{fig:data_collection}: \textit{identification}, \textit{\{Context / Answer\} collection}, and \textit{validation}. Additional data collection details with interface screenshots can be found in Appendix~\ref{sec:appendix_data_collection}.

\subsection{Identification}
We source questions from a variety of datasets for open-retrieval QA: Natural Questions (NQ-Open)~\cite{Kwiatkowski2019NaturalQA}, WebQuestions~\cite{Berant2013SemanticPO}, TyDi-QA~\cite{Clark2020TyDiQA}, and MS-MARCO~\cite{Campos2016MSMA}. All of these datasets are in English and contain questions that are answerable by Wikipedia documents.

Temporally dependent questions are abundant in existing datasets, as they are unambiguous during annotation and annotators simply provide the current answer at the time of collection. In contrast, geographically dependent questions are often rejected during annotation due to their ambiguity, as there is no consensus among annotators on the assumed geographical context. To correct for this scarcity, we \textit{generate} geographically dependent questions by modifying existing questions from NQ-Open. We identify questions with phrases that specify a location by running an NER tagger \cite{Peters2017SemisupervisedST} and remove the location entity using heuristics based on its syntactic role as identified by a dependency parser \cite{Dozat2017DeepBA}. See Appendix~\ref{sec:appendix_data_collection} for full details.

For each question, we collect 3-way yes / maybe / no annotations for whether they are temporally or geographically dependent. We label questions as context dependent if at least two of annotators label so. We discard questions with split annotations and majority maybe annotations, and map the labels to a binary yes/no label.\footnote{These usually are highly ambiguous questions, such as ``What are the measurements of a vintage oil can?'', and total discarded questions only accounts for 11\% of the data.}

\subsection{\{Context / Answer\} Collection}\label{sec:ctx_ans_collection}
After identifying context-dependent questions, we then move on to collecting multiple \{Context / Answer\} pairs via crowdsourcing. For this part, we exclusively use questions from NQ-Open dataset. We allow crowdworkers to query an up-to-date version of English Wikipedia using the Google Search API.\footnote{We use the February 20, 2021 dump of English Wikipedia.} We outline the annotation process for each context type below:

\paragraph{\textsc{temp}} 
To construct \situatedqa~ examples with temporal context, we take two steps: (1) crowd sourcing timeline for each question and (2) generating $(q, c_v, a)$ from the annotated timeline.

Crowdworkers are asked to provide a brief timeline of answers for a given question, consisting of the current answer, the previous answer, as well as the start and end transition timestamps for each answer given as dates or years (see Figure~\ref{fig:data_collection}). From annotated timeline and query pairs, we construct examples of $(q, c_v, a)$ pairs in one of three ways which we refer to as \textbf{Start}, \textbf{Sampled}, and \textbf{Static}. We describe each of these methods below:

\textbf{Start} examples simply use each answer's start transition timestamp as $c_v$. This is intended to simulate the common scenario of asking information that is new or has recently changed. \textbf{Sampled} examples use a timestamp that lies between an answer's start and end transition as $c_v$. As each answer in a timeline can result in many valid $(q, c_v, a)$ pairs, we limit the number of values of $c_v$ for each answer to a maximum of two. We uniformly sample up to two timestamps between each answers start and end transition, using each sampled timestamp to create a new $(q, c_v, a)$ triple. \textbf{Static} examples utilize questions that were annotated as \textbf{not} temporally-dependent to simulate realistic settings where context dependent and independent questions coexist. For each static question, we uniformly sample a single value of $c_v$, resulting in one $(q, c_v, a)$ triple per static question.



\paragraph{\textsc{geo}}
We construct our evaluation datasets with crowdsourcing. For each stripped question that was annotated as geographically dependent during the identification stage, annotators were presented with the question, the original answer(s), and stripped location. Annotators first validate whether the original location and answer pair is correct for the stripped question before identifying up to two additional location and answer pairs. We then use each validated or identified pair to construct a $(q, c_v, a)$ triple for the stripped question. 

We construct our training set via distant supervision. Using our best performing context-dependent question identification model which we introduce in Section~\ref{sec:baselines}, we classify unlabeled examples from the NQ-Open training set. For each example that was classified as geographically-dependent, we use its original answer, location, and stripped question to construct a $(q, c_v, a)$ triple.

\begin{table*}
\footnotesize
\begin{center}
\begin{tabular}{l|rr|rr|rr||r|rrr||r}
\toprule
& \multicolumn{7}{c|}{Identification Data} & \multicolumn{4}{c}{\{Context / Answer\} Data} \\
\multicolumn{1}{l|}{Context}
    & \multicolumn{2}{c|}{Train} & \multicolumn{2}{c|}{Dev} & \multicolumn{2}{c||}{Test} & \multicolumn{1}{c|}{Agreement}
    & \multicolumn{1}{c}{Train}  & \multicolumn{1}{c}{Dev}  & \multicolumn{1}{c||}{Test} & \multicolumn{1}{c}{Agreement} \\ 
\multicolumn{1}{l|}{Type ($c_t$)} & \# $q$ & \% & \# $q$ & \% & \# $q$ & \% & Fleiss' $\kappa$ & \# $q, c_v, a$ & \# $q, c_v, a$ & \# $q, c_v, a$ & Fleiss' $\kappa$ \\ \midrule
\textsc{temp} & 4438 &36 &2572 &32 &1962 & 28& 0.62& 6009 & 3423 & 2795 &0.57 \\
\textsc{geo} & 1149 & 46 & 879 & 42 & 367 & 37 & 0.56 & 3548 & 1398 & 505 & 0.56 \\
\bottomrule
\end{tabular}
\end{center}
\caption{Dataset statistics. For our identification dataset, we report both the total number of questions (\# $q$) and the percent of questions that are context-dependent (\%). For our \{Context / Answer\} dataset, we report the number of unique (question, context value, answer) triples (\#~$q, c_v, a$). Each context type and split's $(q, c_v, a)$ triples are collected slightly differently, see Section~\ref{sec:ctx_ans_collection} for details. We also report inter-annotator agreement on context dependent question identification and \{Context / Answer\} validation.}
\vspace{-0.8em}
\label{tab:data_stats}
\end{table*}

\subsection{Quality Control}\label{sec:quality_control}
\paragraph{Validation}
The \{Context / Answer\} annotations collected above are reviewed by another annotator in a final validation stage. Presented with a question along with all answer timelines or location/answer pairs that have been generated by other workers, annotators are asked to validate or revise each response by marking them as correct or incorrect. We collect 2-way \{Context / Answer\} annotations and have one validatator for each question from the development and test sets. We collect a single \{Context / Answer\} annotation and skip validation for answer timelines from the training set.

\paragraph{Worker Qualifications / Inter-annotator agreement}
Table~\ref{tab:data_stats} reports Fleiss's kappa for the context dependent question identification and validation stages. We find moderate to high agreement all tasks, with lower agreement on datasets with highly ambiguous queries (e.g. how much to trust friends) or questions require extensive domain knowledge (e.g., what is the cause of smog in china?). For geographically dependent question answering, questions like ``when were electric trains introduced" had split votes from the annotators, as both geographically dependent and independent interpretations are valid. Overall, we observe high quality data with comparable agreement numbers with prior work, yet reaffirming prevalent ambiguity in open retrieval QA~\cite{Min2021NeurIPS2E}.

\subsection{Data Statistics / Analysis}\label{sec:data_analysis}
Table~\ref{tab:data_stats} shows the statistics of our collected dataset. We annotated over 11K questions, and roughly 30-40\% of them where identified as context-dependent. Temporally-dependent questions comprised of least 10\% of examples in all datasets we looked at, even without any filtering. We also found that examples from Natural Questions where at least one answer span is within a table in its evidence document are more often temporally-dependent. To construct our final dataset, we upsample such questions. We followed the original train, development, and test split for each dataset, which caused slight inconsistencies in the proportion of examples from each dataset across splits, simulating a domain shift. 

For 2.8K of those context-dependent questions (2.4K temporal, 0.5K geographical), we collected a total of 5.9K answers from alternate contexts (4.0K temporal, 1.9K geographical). From those alternate temporal context / answer pairs, we construct ($q, c_v, a$) by sampling valid dates, creating 6K examples. The final \textsc{temp} dataset also includes 6.7K examples from temporally-independent questions. More details and statistics on individual datasets can be found in the in Appendix~\ref{sec:appendix_data_collection}.

\begin{figure}
    \centering
    \includegraphics{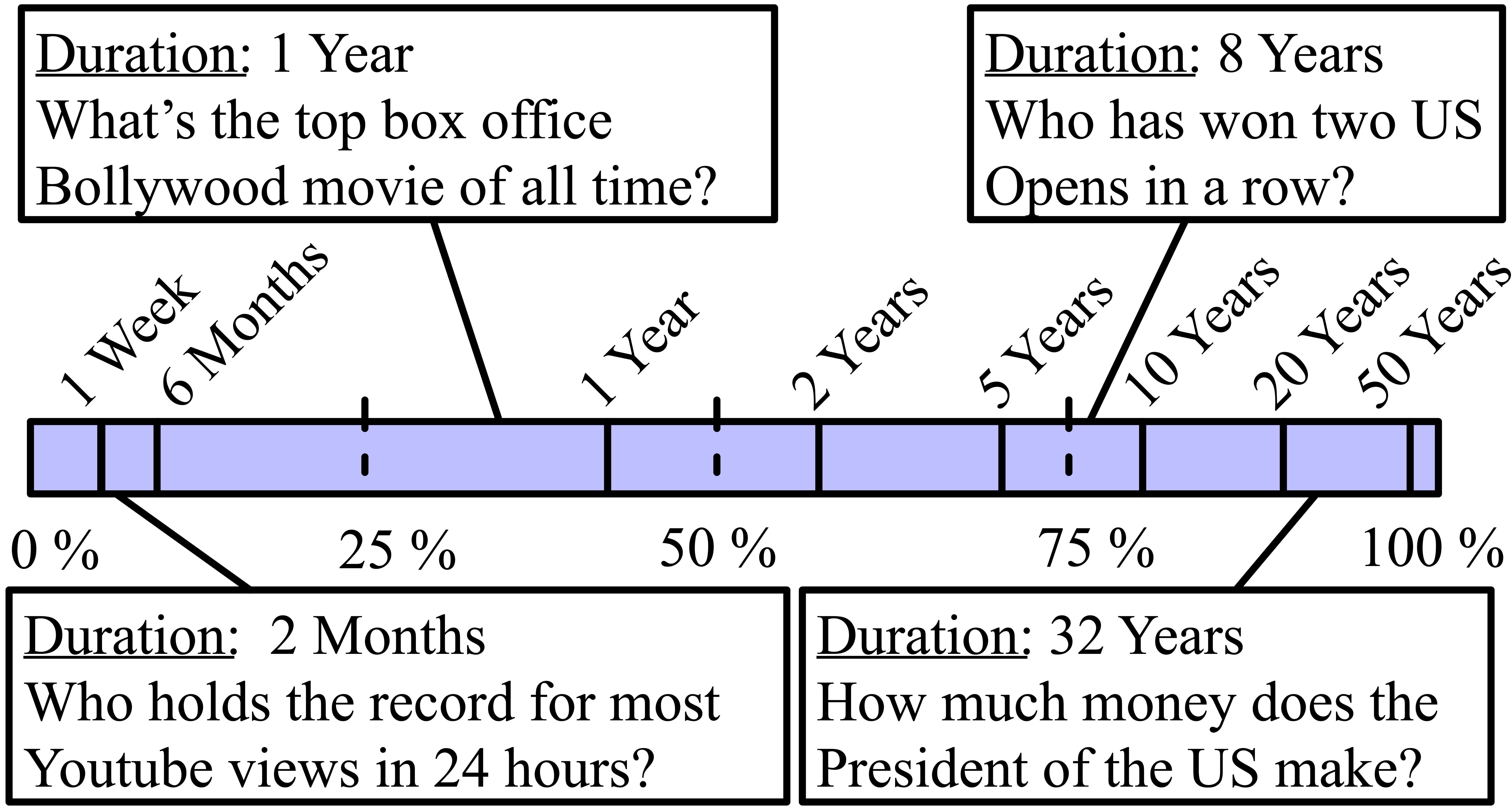}
    \caption{The distribution of temporally dependent questions by the duration its previous answer was true for with examples.}
    \label{fig:duration}
    \vspace{-0.8em}
\end{figure}

\paragraph{How often do temporally dependent facts change?}
We investigate how frequently answers change by measuring the distance between the the two most recent answers. Figure~\ref{fig:duration} depicts how long the previous answer was valid for. We observe a long tailed distribution, with a large proportion of answers changing around the one year mark. 

\section{Models}\label{sec:baselines}
In this section, we describe our baseline models for context-dependent question identification and situated QA. Details on our learning settings and hyperparameters are in Appendix~\ref{sec:appendix_implementation}.

\subsection{Context Dependent Question Identification}
As a lower bound, we provide a naive random-choice baseline, where labels are randomly assigned to match the label distribution in the training data. We also train separate BERT-based classifiers, which encode the input question $q$ using BERT and classify based on each question's encoded \textsc{[CLS]} token representation. We experiment with both BERT-base and BERT-large. We approximate human performance by sampling 50 examples for temporal context and geographical context each, and re-collecting annotations from a new set of crowdworkers.

\subsection{Situated Question Answering}
There are two types of models for open retrieval QA: retrieval based (open book) and closed book. Retrieval based models~\cite{karpukhin2020dense,lee-etal-2019-latent,guu2020realm} answer questions by retrieving relevant passages from a corpus before producing an answer based on the retrieved passages. Closed book approaches~\cite{2020t5cqba,lewis-etal-2020-bart,Fvry2020EntitiesAE} do not rely on access to a corpus during inference, instead use the knowledge that has been stored in the model's parameters. These models, which typically have a large number of parameters and are trained over a single snapshot of web data, generate answers from each question.

We present two competitive baselines: one retrieval-based model, DPR \cite{karpukhin2020dense}, and one closed-book model, BART-large \cite{lewis-etal-2020-bart}.\footnote{We use the pretrained model and implementation from \url{https://github.com/shmsw25/bart-closed-book-qa}.} Both baselines are trained on NQ-Open~\cite{Kwiatkowski2019NaturalQA,lee-etal-2019-latent}. We use the best performing model configuration from DPR paper, but swap the retrieval corpus with an up-to-date Wikipedia dump (timestamped 2021-Feb-20). In standard settings, DPR achieves an accuracy of 41.5 and BART achieves an accuracy of 24.1 on NQ-Open. We use these baselines in three settings which we introduce below: unmodified, query modified, and query modified with finetuning.

We also approximate human performance on situated QA using 100 randomly sampled examples from each context type which are annotated by the authors of this paper. While our estimated human performance appear somewhat low, they are in line with agreement rates in the original Natural Questions study~\cite{Kwiatkowski2019NaturalQA} and are tied to the challenges of evaluating open-retrieval QA. Discrepancies are often due to ambiguous questions or equivalent answer forms (e.g., July 5, 1945 vs. 5 July 1945, The Patriots vs. New England Patriots). The latter is especially problematic for geographically dependent questions, where answers involving time expression occur much more frequently, comprising over 50\% of examples. See Appendix~\ref{sec:appendix_data_collection} for more details.

\begin{table}
\footnotesize
\begin{center}
\begin{tabular}{p{2.5cm} p{4.25cm}}
\toprule
Context $(c_t, c_v)$ & Augmented Question \\ \midrule
$(\textsc{temp}, \textsc{2017})$ & Who went number 1 in the WNBA draft \textbf{as of 2017}?  \\ 
$(\textsc{geo}, \textsc{Tokyo})$ & Where do we rank among the world’s largest cities \textbf{in Tokyo}? \\
\bottomrule
\end{tabular}
\end{center}
\caption{Examples of modified queries. Text shown in bold is concatenated onto the original question. To query for the answer for some context where $c_t=\textsc{temp}$ or $c_t=\textsc{geo}$, we append the phrase ``as of $c_v$'' or ``in $c_v$'', replacing $c_v$ with the context value.}
\vspace{-0.8em}
\label{tab:ctx_augmentation}
\end{table}

\paragraph{Unmodified Baseline}
We evaluate open-retrieval QA models on our task without any alterations. Each model is only given the question, ignoring the extra-linguistic context, and thus predicts the same answer no matter the context. For temporally-dependent questions, models are expected to output the most up-to-date answer. For geographically-dependent questions, models must assume some geographical context to produce an answer.

\paragraph{Query Modified}
One simple method for incorporating extra-linguistic contexts into question answering is to concatenate the context onto the question, separated by some special token. We adopt a slightly different approach that leverages what models have already learned about producing answers from specific contexts: concatenating a phrase that specifies the relevant context onto each question, transforming it into a context-independent question. The concatenation templates for each context type are described in Table~\ref{tab:ctx_augmentation}. We find that these simple modifications mostly generate valid, fluent questions that closely resemble examples found in our models' training data. We estimate that 10\% of questions in NQ-Open contain similar augmentations, inserting or concatenating a phrase that specifies the context (See Appendix~\ref{sec:appendix_data_analysis} for details).

\paragraph{Query Modified w/ Finetuning}
We also experiment with finetuning our query modified baselines. For our retrieval based approach (DPR), we finetune separate the reader and retriever models for each context type. We finetune retriever models using gold and hard-negative passages from the retrieval results of our Query Modified DPR baseline. Likewise, we also finetune separate closed book models for each context type.

\begin{table*}[]
\footnotesize
\begin{center}
\begin{tabular}{l|rrrr|rrrr}
\toprule
& \multicolumn{4}{c|}{\textsc{temp}} & \multicolumn{4}{c}{\textsc{geo}}\\
Model & Accuracy & Precision & Recall & F1  & Accuracy & Precision & Recall & F1 \\ \midrule
Random & 45.3 / 44.8 & 31.9 / 28.3 & 61.7 / 64.9 & 42.1 / 39.4 & 49.7 / 48.0 & 42.3 / 38.1 & 54.2 / 62.8 & 47.5 / 47.4 \\
$\text{BERT}_\text{base}$ & 94.5 / 93.5 & 94.2 / 92.7 & 88.5 / 83.1 & 91.2 / 87.6 & 86.0 / 79.6 & 79.0 / 67.4 & 90.8 / 87.6 & 84.5 / 76.2 \\
$\text{BERT}_\text{large}$ & 95.7 / 93.8 & 94.7 / 91.2 & 91.7 / 85.9 & 93.2 / 88.5 & 86.1 / 80.7  & 80.3 / 69.9 & 88.6 / 84.7 & 84.3 / 76.6 \\ \midrule
Human$^*$ & 96.0 / 94.0 & 88.9 / 83.3 & 100 / 100 & 94.1 / 90.9 & 88.0 / 84.0 &  76.5 / 72.7 & 86.7 / 88.9 &  81.2 / 80.0 \\
\bottomrule
\end{tabular} 
\end{center}
\caption{Results for context-dependent question identification. Each cell report results on development / test set.}
\vspace{-0.8em}
\label{tab:id_results}
\end{table*}

\begin{table*}
\small
\begin{center}
\begin{tabular}{lcc|rrrr|rrr}
\toprule
&\multirow{2}{2em}{Query Mod.} & \multirow{2}{2.2em}{Fine-tuned} &  \multicolumn{4}{c|}{\textsc{temp}} & \multicolumn{3}{c}{\textsc{geo}} \\
&  &  & Static (400) & Samp. (1472) & Start (923) & Total  & Comm. (265) & Rare (240) & Total \\\midrule
\multirow{3}{3em}{Retrieval based} &&&  \textbf{44.2} & 16.0 & 14.2 & 19.4 &  9.1 &  2.9 &  6.1 \\
& \checkmark &&                         28.8 & 15.9 & 18.5 & 18.6 & 27.5 & 22.1 & 25.0 \\
& \checkmark & \checkmark &             39.8 & \textbf{17.2} & \textbf{24.9} & \textbf{23.0} & \textbf{27.9} &\textbf{ 25.0} & \textbf{26.5} \\ \midrule
\multirow{3}{3em}{Closed Book} &&&      27.2 & 15.3 & 12.9 & 16.2 &  9.4 &  4.6 &  7.1 \\
& \checkmark &&                         19.5 & 12.4 & 15.7 & 14.5 & 19.2 &  9.2 & 14.5 \\
& \checkmark & \checkmark &             26.0 & 16.2 & 18.3 & 18.3 & 21.5 & 11.7 & 16.8 \\\midrule
Human$^*$ &&&                           - & - & - & 57.0 & - & - & 34.0 \\
\bottomrule
\end{tabular} 
\end{center}
\caption{Results situated question answering, reporting exact match score on the test set. In addition to reporting overall EM for each context type, we also report EM for partitions of the test set. For \textsc{temp}, we partition the test set based on how the example's context value was generated. For \textsc{geo}, we split on whether the context-value is a location is common (Comm.) or uncommon (Rare), which is determined by whether the location appears at least five times in our dataset as a geographical context.}
\label{tab:qa_results}
\end{table*}

\section{Results}\label{sec:results}

\subsection{Context Dependent Question Identification}
Table~\ref{tab:id_results} reports our results on context dependent question identification. We find that pretrained language models perform competitively in this binary classification task, matching human agreements. Thus, it may be a useful tool for identifying context dependent questions which closed book systems are poorly suited for or identifying examples in benchmark datasets that require re-annotation. 

\subsection{Situated Question Answering}
Table~\ref{tab:qa_results} shows our results on the situated QA task. All models lag behind our estimated human-level performance by a significant margin, especially for temporally dependent questions. Overall, retrieval based models outperform closed book models, as in the original QA datasets (41.5 EM vs. 24.1 EM). We find that applying our query modifications to the question works well for geographically dependent questions, but see no improvement for temporally dependent questions. Finetuning models on answers from multiple extra-linguistic contexts with modified questions improved performance across the board, especially for the \textsc{temp} context type.

Adding \textsc{temp} context to static questions distracts the model, significantly decreasing performance especially for the closed book baseline. This suggests that models are not robust to semantically equivalent edits, as observed in~\citet{Ribeiro2020BeyondAB}. Models also perform better when provided with an answer's start date as context compared to a sampled date, which falls between the transitions between answers. This gap is especially pronounced for our retrieval based model.

\begin{table}
\footnotesize
\begin{center}
\begin{tabular}{lcc|rr|rr}
\toprule
&\multirow{2}{2.2em}{Query Mod.} & \multirow{2}{2.2em}{Fine-tuned} &  \multicolumn{2}{c|}{\textsc{temp}} & \multicolumn{2}{c}{\textsc{geo}} \\
&  &  &\textit{One} & \textit{Any} & \textit{One} & \textit{Any} \\ \midrule
\multirow{3}{2.75em}{Retrieval based} &&&  19.4 & 28.1 &  6.1 & 22.6 \\
& \checkmark &&                            18.6 & 26.4 & 25.0 & 30.7 \\
& \checkmark & \checkmark &                \textbf{23.0} & \textbf{29.7} & \textbf{26.5} & \textbf{32.3} \\ \midrule
\multirow{3}{2.75em}{Closed Book} &&&      16.2 & 24.0 &  7.1 & 25.1 \\
& \checkmark &&                            14.5 & 19.9 & 14.5 & 22.6 \\
& \checkmark & \checkmark &                18.3 & 25.9 & 16.8 & 24.4 \\
\bottomrule
\end{tabular} 
\end{center}

\caption{Breakdown of errors from context dependent QA. We report EM accuracy evaluated against answers from the  correct context (\textit{One}) and answers from the union of all our annotated contexts (\textit{Any}).}
\label{tab:qa_error_analysis}
\end{table}


\paragraph{Do systems performance vary based on the frequency of geographic context?}
Table~\ref{tab:qa_results} splits results on \textsc{geo} questions into those from common and rare locations, many of which unseen during training. Both methods have greater difficulty with rare locations; however, the drop in performance for closed book models is significantly worse. Closed book models perform 10 percentage points worse on questions from rare locations while retrieval based models only perform 3 points worse. These findings show that retrieval based systems generalize better to new locations and are better equipped for answering geographically dependent questions. We also see that the gap in performance between common and rare locations shrinks for both models after finetuning, suggesting that explicitly modeling geographical contexts helps with generalization.

\begin{table}[]
\footnotesize
\centering
\begin{tabular}{l|rrrr}
    \toprule
                   & US & India & Other & Rare \\
    \# of Examples & 42 &    20 &   203 &  240 \\ \midrule
    Retrieval Based        & 23.8 & 10.0 & 5.9 & 2.9 \\
    Closed Book            & 23.8 & 10.0 & 6.4 & 4.6 \\
    \bottomrule
\end{tabular}
\caption{Results of our unmodified baseline run on \textsc{geo}. We further break down performance common locations by separating the two most common ones, the ``United States'' and ``India'',  the ``Other'' common locations.}
\label{tab:location_bias}
\end{table}

\begin{table}
\footnotesize
\begin{center}
\begin{tabular}{lcc|rrr}
\toprule
& \multirow{2}{2.2em}{Query Mod.} & \multirow{2}{2.2em}{Fine-tuned} &  Current & Previous & Total \\
&&& 529 & 373 & 902 \\ \midrule
\multirow{3}{2.75em}{Retrieval based} &&& 19.5 &  6.2 & 14.0 \\
& \checkmark &&                           19.5 &  6.4 & 14.1 \\
& \checkmark & \checkmark &               \textbf{22.5} &\textbf{ 11.3} &\textbf{ 17.8} \\ \midrule
\multirow{3}{2.75em}{Closed Book} &&&     18.5 &  4.0 & 12.5 \\
& \checkmark &&                           18.5 &  4.8 & 12.9 \\
& \checkmark & \checkmark &               18.9 &  8.0 & 14.4 \\ \midrule
Human$^*$ &&& - & - & 51.0 \\
\bottomrule
\end{tabular} 
\end{center}
\vspace{-0.7em}
\caption{Results on training our baselines on predicting the current and previous answers to a question. Human performance is estimated using 100 randomly sampled examples annotated by the authors of this paper.}
\vspace{-0.7em}
\label{tab:temp_rel_results}
\end{table}

\paragraph{What is the assumed context for a geographically dependent question?}
When presented with a geographically dependent question without the context, open retrieval QA models must assume some geographical context to produce an answer. We hypothesize that models will exhibit bias toward assuming the few geographical co-texts are most frequently asked about. We test this by reporting the results of our unmodified baseline, further breaking down the results from Table~\ref{tab:qa_results} by individual location. Each time the unmodified baseline correctly answers a question for some location, it did so by assuming that location as its context. We report our results in In Table~\ref{tab:location_bias}, which show that models are heavily biased toward assuming the question was asked by someone in the United States or India.

\paragraph{How often do models provide answers from the wrong context?}
In Table~\ref{tab:qa_error_analysis}, we report error analysis for our situated QA baselines that shows how often models are provide the answer from the specified context versus the union of all annotated contexts. We see that models often fail to incorporate extra-linguistic contexts into the question, producing the answer from another context. 


\section{Analysis}\label{sec:analysis}

\paragraph{Can QA systems differentiate and recall previous answers to a question?}
To answer this, we study a new setting where we query models for the current and previous answer to a given question. Following the same steps as above for the \textsc{temp} and \textsc{geo} context types, we develop a suite of retrieval-based and closed-book baselines for this new setting. Our query-modified baselines in this setting consist of prepending the word ``previously'' to the beginning of each question to query for the previous answer. In the existing framing of QA, all questions are looking for the current answer we do no augmentations to query for the current answer. For our finetuned baselines, we train separate reader models and closed book models to query for current and previous answers. 

We report our results for these experiments in Table~\ref{tab:temp_rel_results}. We find that models are far better at producing the current answer versus past answers to a question. Finetuning, however, greatly increases performance on queries for previous answers.

\begin{table}[]
\footnotesize
\centering
\begin{tabular}{lllrr}
    \toprule
    & Fine- & Knowledge & Stable & Updated \\ 
    & tuned & Source Date & 267     &  209 \\ \midrule
    \multirow{3}{2.75em}{Retrieval Based} &  & Wiki. Dec 2018 & 27.3 &  7.2 \\
    &                                        & Wiki. Feb 2021 & 26.6 & 11.5 \\
    & \checkmark                             & Wiki. Feb 2021 & 28.1 & \textbf{17.2} \\\midrule
    \multirow{2}{2.75em}{Closed Book} &      & Feb 2019         & \textbf{28.8} &  5.3 \\
    & \checkmark                             & Feb 2019         & 28.1 &  6.7 \\
    \bottomrule
\end{tabular}
\vspace{-0.4em}
\caption{Results on the default open-retrieval QA setting where all questions are assumed to be about the present timestamp (Feb 2021), evaluation split by questions that require updating the world state to \textit{after} the large scale training dataset was collected, \textit{Updated}, and questions where answers didn't change since the data collection, \textit{Stable}. Our retrieval based systems use the English Wikipedia dumps from 2018-12-20 and 2021-02-20 while our closed book systems are pretrained on data from 2019-02 and earlier.}
\label{tab:corpora_update}
\end{table}

\paragraph{Can QA systems trained on outdated answers adapt to the present?}
We investigate whether QA systems that are pretrained on outdated corpora and that are trained on large-scale QA datasets containing outdated answers can adapt to answering questions situated in the present. \citet{Lewis2020RetrievalAugmentedGF} shows that updating the corpera of retrieval based models can be effective method for updating simple facts, such as current heads of state. We test whether this extends to a broader range of temporally-dependent facts by evaluating our baselines introduced above on their ability to predict the current answer in our dataset, which are up-to-date as of Feb 2021. We split the current answers into two categories, \textit{stable} where answer did not change since 2018, and \textit{updated} where answer changed after 2018. We chose 2018 as a threshold as it is when Natural Questions~\cite{Kwiatkowski2019NaturalQA} was collected and roughly matches the timestamp of the most recent data our closed book model~\cite{lewis-etal-2020-bart} was pretrained on (Feb 2019).

Table~\ref{tab:corpora_update} shows that, while retrieval based models were able to update some world knowledge after swapping the retrieval corpus, significant gains on \textit{updated} questions only come after finetuning with newer data. This suggests that simply updating the corpora that models retrieve passages from is not sufficient to keep models up-to-date.

\section{Related Work}

\paragraph{Comparison to other QA datasets}

While a few prior studies mention that answers to a question can change over time~\cite{min2020ambigqa}, no prior work investigated the performance of existing models on temporally dependent queries and their prevalence.

Addressing ambiguous questions~\cite{min2020ambigqa} overlaps with our study, as ambiguity can arise from extra-linguistic contexts. Their study found 13\% of examples were ambiguous due to temporal deixis. However, we found that context-dependence can co-occur with inherent ambiguities in the question which cannot be resolved with context. Thus, this figure underestimates the true proportion of temporally dependent questions. Furthermore, humans don't always consider context-dependent questions as ambiguous when there is only one answer given the context, which is normally assumed to be present time and location. Thus, ambiguities that arise due to lack of context should, therefore, be modeled separately from semantic ambiguities of the question.

Recent work has also studied generating temporally-dependent questions with timestamped answers from temporal knowledge-bases~\cite{Chen2021ADF,Saxena2021QuestionAO}. Similarly,~\citet{Dhingra2021TimeAwareLM} generates temporally-dependent cloze-style prompts and presents a temporally aware language model to address them. These works, however, synthetically generate examples and are therefore limited in scope and diversity. In contrast, we manually annotate temporally and geographically dependent questions.




\paragraph{Temporal understanding}

Temporal understanding in NLP has been typically studied in an intra-document setting, i.e., ordering events or finding temporal relationship between two events in the same document~\cite{Pustejovsky2003TimeMLRS,Bethard2007TimelinesFT,Cassidy2014AnAF,Llorens2015SemEval2015T5,Zhou2020TemporalCS}.

\paragraph{Dynamic evaluation}
Dynamic evaluation has been studied under the language modeling~\cite{osborne2014exponential,Yogatama2014DynamicLM}, topic modeling~\cite{Wang2008ContinuousTD} and entity linking~\cite{Rijhwani2020TemporallyInformedAO}. Recent work~\cite{Lazaridou2021PitfallsOS} studied temporal drift in the context of large scale pretrained language model, hinting reusing models from previous snapshot can cause performance decay, as we observe in our study as well. 

Adversarial, phased data collection has been proposed~\cite{Paperno2016TheLD,Zellers2018SWAGAL,Potts2020DynaSentAD} to drive model development, constantly feeding models examples that current models are unable to address. We suggest collecting questions dynamically for slightly different goal, to accurately reflect the changing world by identifying temporally dependent facts from such benchmarks and continuously updating them. Building a benchmark based on a fixed snapshot~\cite{Petroni2020KILTAB} can be a viable alternative.

Researchers studied keeping knowledge sources up-to-date. \citet{Konovalov2017LearningTE} looked at the constantly changing facts from the perspective of automatically extracting knowledgebase revisions. ~\citet{Schuster2021GetYV} presents contrastive sentences based on Wikipedia revisions, showing how entailment decision can change over time.

\section{Conclusion \& Future Work}\label{sec:future_work}
We present the first study of how extra-linguistic contexts affect open retrieval QA. Our study reveals that current systems fail to adapt to shifts in the temporal or geographical context. We, therefore, propose tasks and create a dataset for training and evaluating QA systems on modeling how facts change across contexts. Our dataset will support ample future work for developing models which can gracefully update its predictions based on new temporal and geographical contexts. 
Future research may address incorporating in temporally and geographically dependent source documents, such as news articles, or considering other extra-linguistic contexts such as \textit{who} is asking the question, taking individual's preferences into account.

\section*{Acknowledgement}
We would like to thank Sewon Min, Raymond Mooney, and members of UT NLP group for comments and discussions. The work is partially funded by Google Faculty Awards. 

\nocite{*}
\bibliography{anthology,custom}
\bibliographystyle{acl_natbib}

\newpage

\section*{Appendix}\label{sec:appendix}

\appendix
\section{Data Collection Details}\label{sec:appendix_data_collection}

\paragraph{Data Composition}
For NaturalQuestions and WebQuestions, we use the open-domain splits established in \cite{lee-etal-2019-latent} and we only consider questions from TyDi-QA that are written in English. We randomly sampled the questions from all datasets except for NQ, where the ratio of temporally dependent questions was low (around 15\%). Initial pilot tests showed that questions where the answer span is within a table in its evidence document (we refer these as NQ-Table) are more often temporally dependent compared to questions where answers are found only in paragraphs (we refer these as NQ-Passage). To better target temporally dependent questions, we sampled heavily from NQ-Table. The source dataset statistics can be found in Table~\ref{tab:ctx_id_data}, and proportion of temporally dependent questions in each dataset can be found in Table~\ref{tab:stage_1_analysis}.

\paragraph{Geographically Dependent Question Generation}
To generate geographically dependent questions, we first run the named entity recognition tagger from \citet{Peters2017SemisupervisedST} and the dependency parser from \citet{Dozat2017DeepBA} over each example and filter for those with a geopolitical entity in the question. We then delete the entity based on its syntactic role. If the entity's syntactic role is either \textit{nn} or \textit{amod}, we delete the entire entity and all of its descendants. If the the entity's role is \textit{pobj}, the entity's parent preposition and all its descendants. Finally, if the entity's role is \textit{root} or \textit{nsubj}, we replace the entity with the pronoun \textit{we}, deleting all determiners and conjugating any auxiliary verbs accordingly. We ignore instances where the dependency is not in one of these categories, there are multiple GPE entities, the stripped questions is has 3 tokens or less, or there is disagreement between our parser and tagger. We use the implementations of \citet{Peters2017SemisupervisedST} and \citet{Dozat2017DeepBA} from AllenNLP \cite{Gardner2017AllenNLP}.

\begin{figure*}
\centering
\includegraphics[width=16cm]{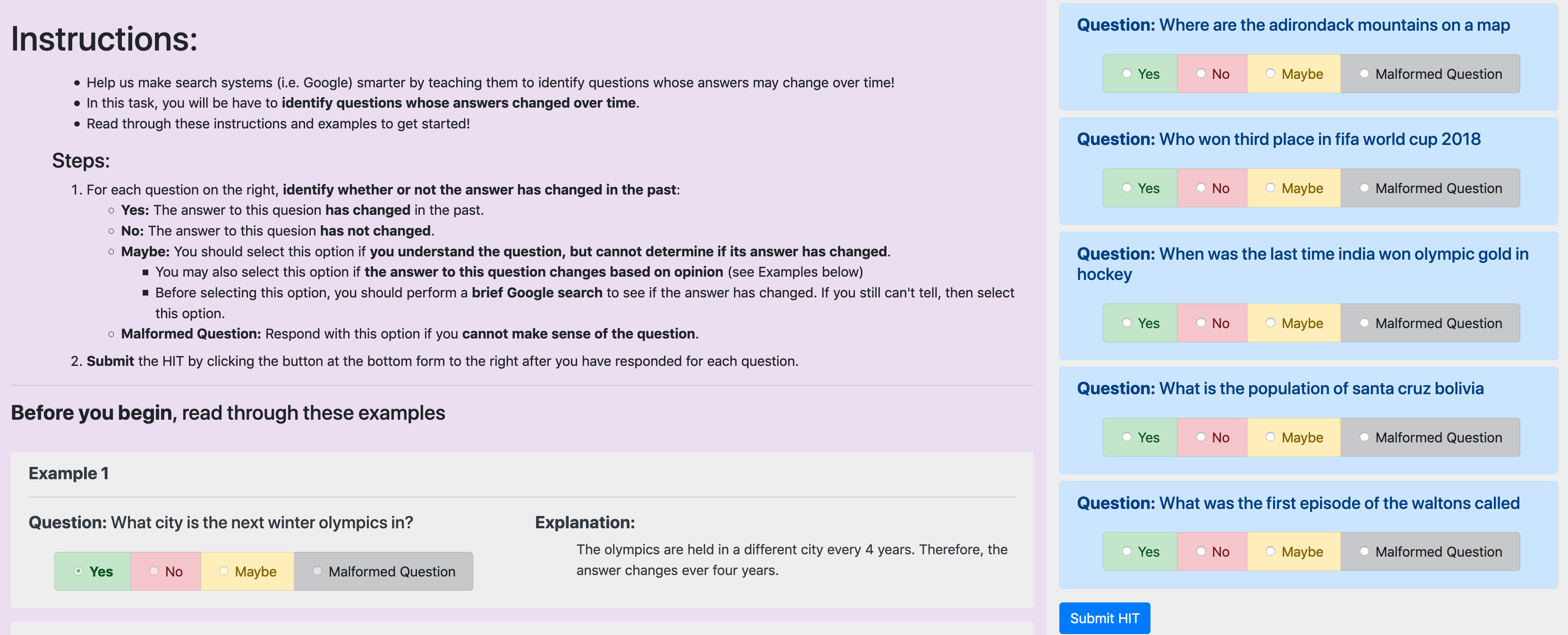}
\caption{Interface for identifying temporally and geographically dependent questions.}
\label{fig:temp_id_interface}
\end{figure*}

\begin{figure*}
\centering
\includegraphics[width=16cm]{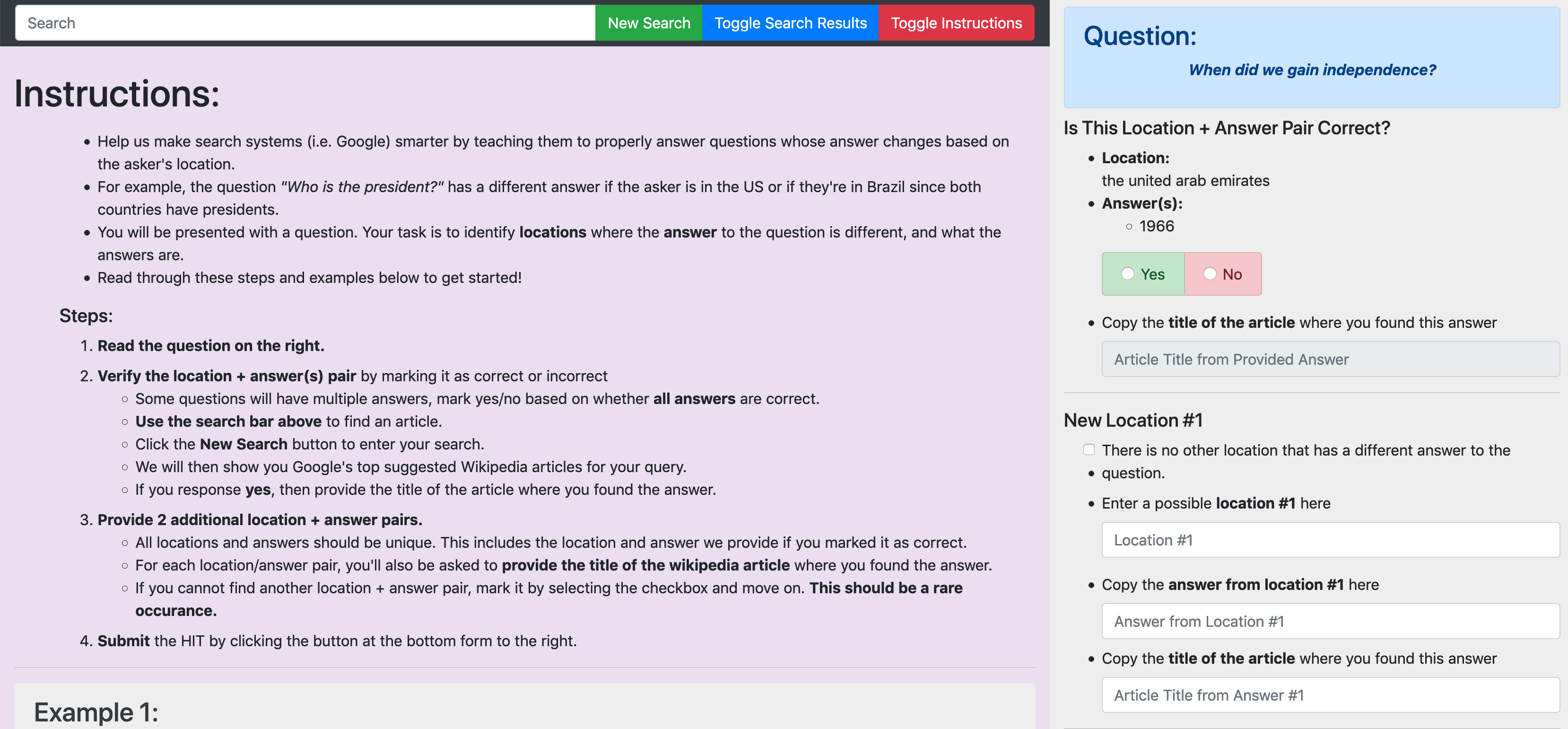}
\caption{Interface for collecting geographical \{Context / Answer\} pairs. Annotators are also asked to verify the original location and answer.}
\label{fig:geo_qa_interface}
\end{figure*}

\begin{figure*}
\centering
\includegraphics[width=16cm]{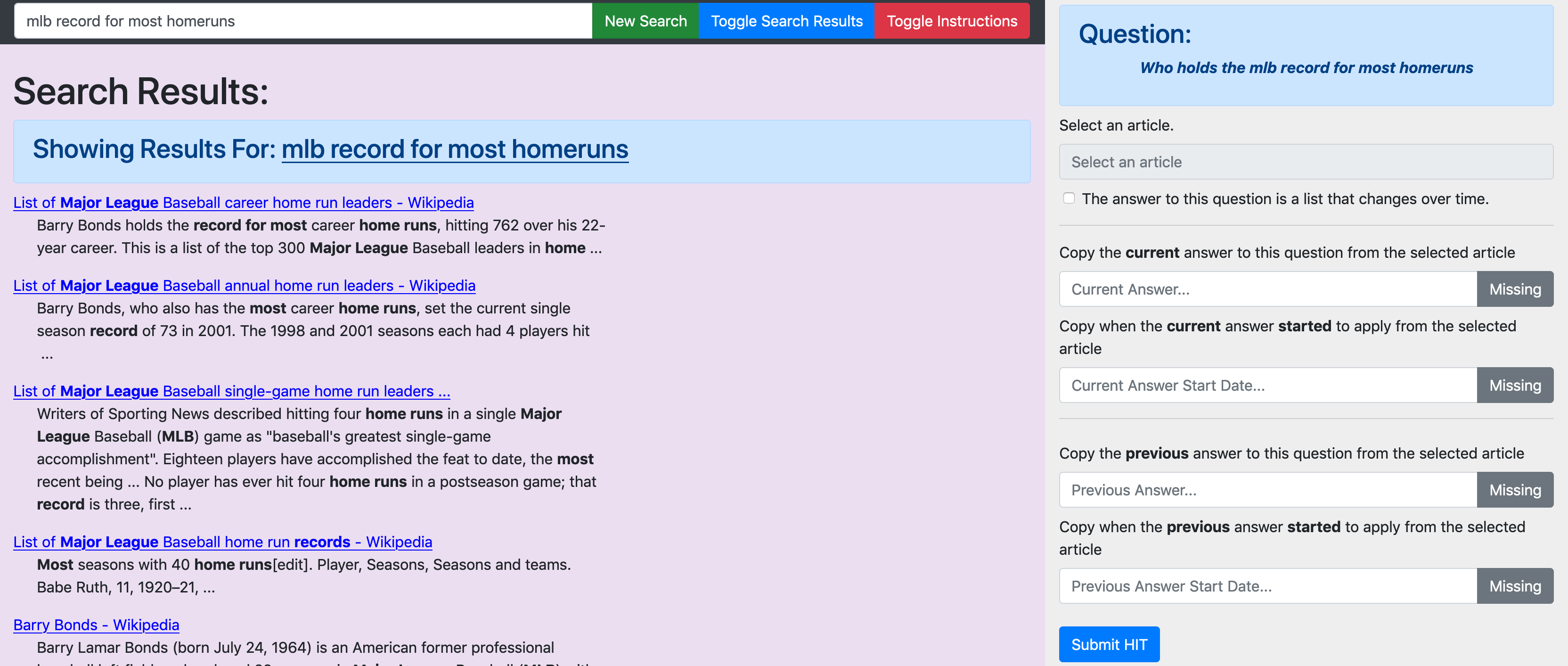}
\caption{Interface for searching for articles during temporal \{Context / Answer\} collection. The search interface is shared between temporal and geographical \{Context / Answer\} collection and verification stages.}
\label{fig:temp_search_interface}
\end{figure*}

\begin{figure*}
\centering
\includegraphics[width=16cm]{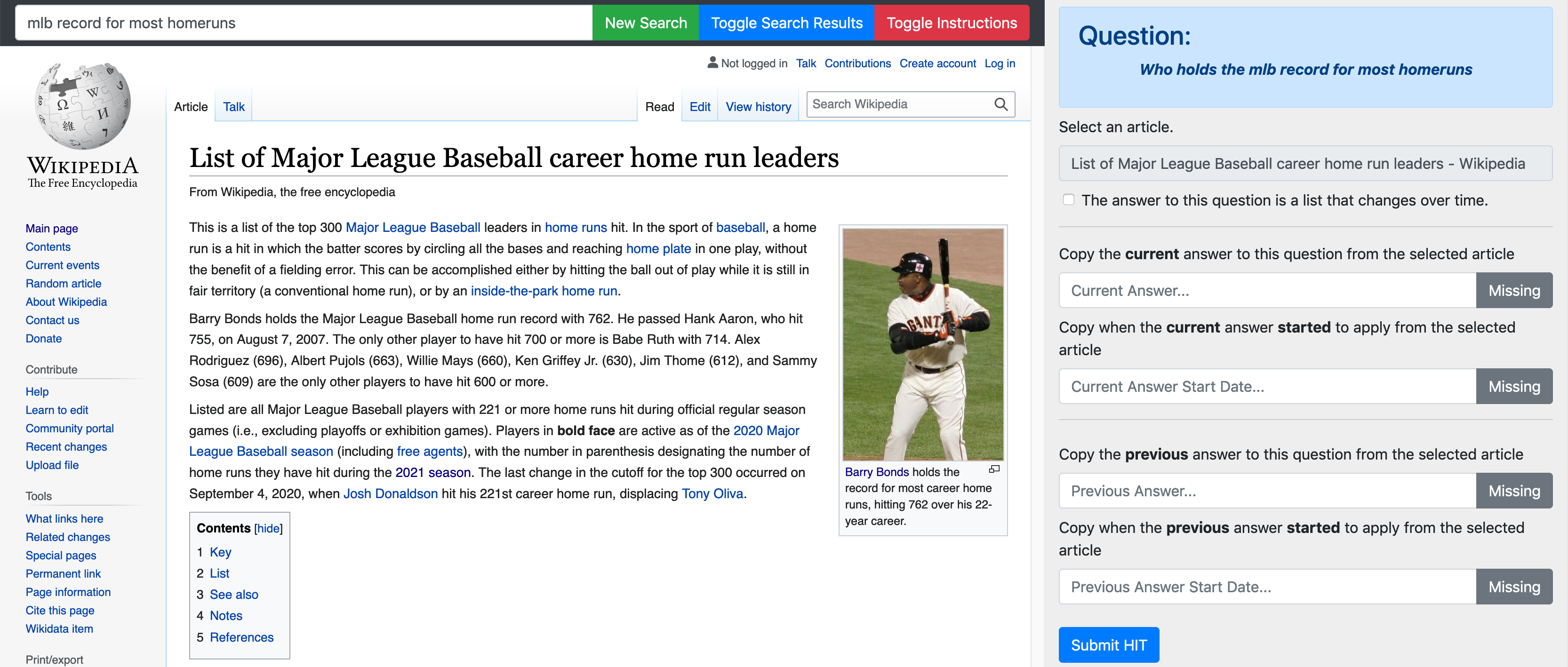}
\caption{Interface for viewing Wikipedia articles during the \{Context / Answer\} collection and verification stages. Workers select articles from the search results depicted above that they want to view.}
\label{fig:temp_wiki_interface}
\end{figure*}

\begin{figure*}
\centering
\includegraphics[width=16cm]{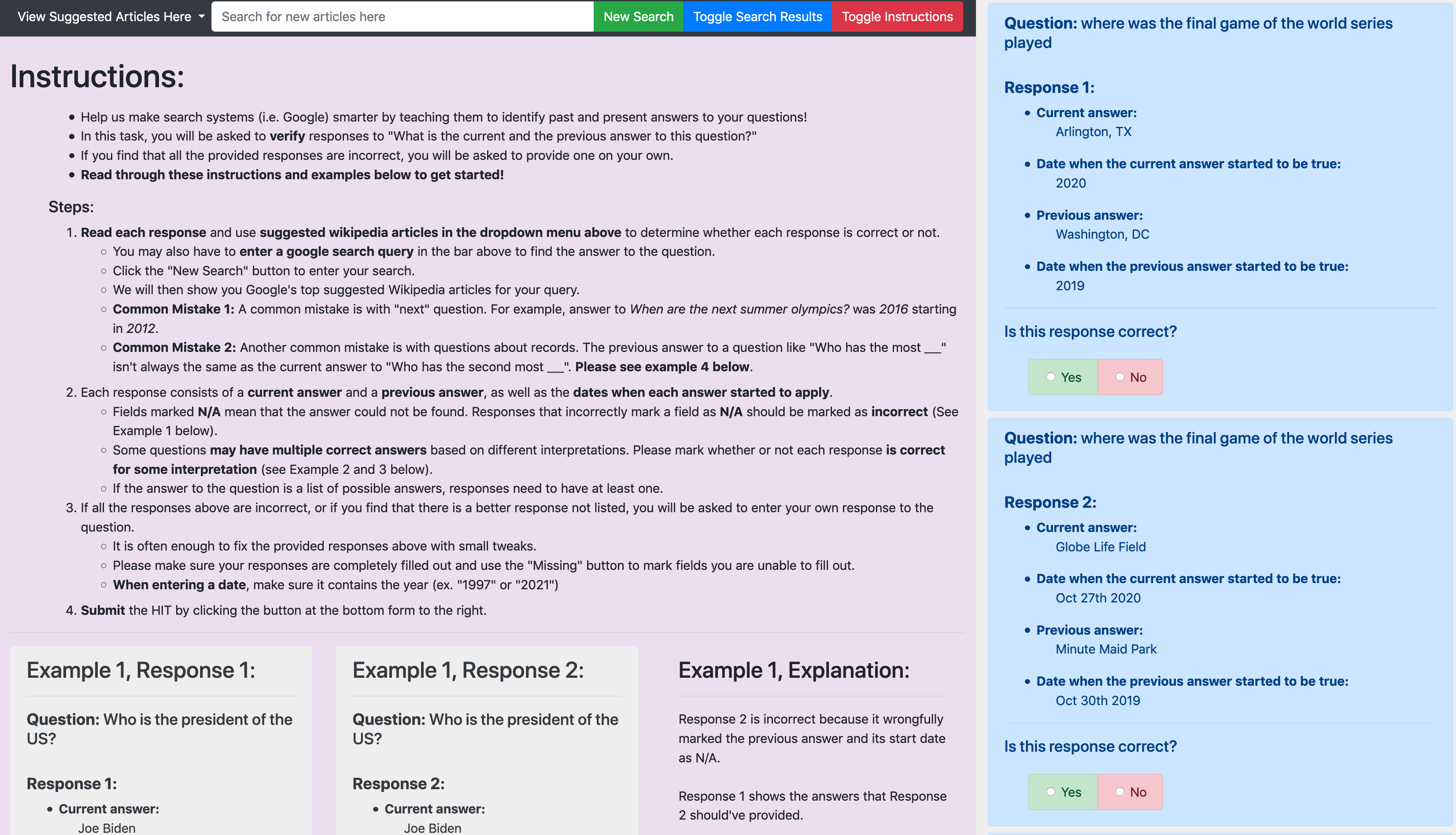}
\caption{Annotation interface for \{Context / Answer\} verification stages. Workers may search for the answers themselves, or view the articles used by workers in the prior stage.}
\label{fig:temp_ver_interface}
\end{figure*}

\paragraph{Crowdsourcing Details}
We pay workers 0.15 USD per identification HIT and 0.40 USD per \{Context / Answer\} collection and validation HIT. During \{Context / Answer\} collection, if an annotator is unable to find the previous answer or either transition date after visiting 3 articles, they may leave out that information. In validation stage, in addition to allowing annotators to search over English Wikipedia, we also provide a list of articles that were used by workers in the prior stage. In open retrieval setting, the same question may have multiple interpretations, and therefore we ask annotators to mark answers as correct if they consistent with one plausible interpretation. Figure~\ref{fig:temp_id_interface},~\ref{fig:geo_qa_interface},~\ref{fig:temp_search_interface},~\ref{fig:temp_wiki_interface},~\ref{fig:temp_ver_interface} shows our annotation interface. 

\begin{table}
\footnotesize
\begin{center}
\begin{tabular}{lrrrr}
\toprule
Source & Train & Dev & Test & Total \\ \midrule
NQ-Table & 2911 & 500 & 839 & 2647 \\
NQ-Passage & 408 & 500 & 500 & 1500 \\
WebQuestions & 367 & 361 & 500  & 861\\
TyDi-QA (en) & 412 & 413 & N/A & 1000 \\
MS-MARCO & 340 & 314 & 348 & 1500 \\ \midrule
Total & 4438 & 2572 & 1962 & 8972 \\
\bottomrule
\end{tabular}
\end{center}
\caption{Number of identification examples by split and source dataset.}
\label{tab:ctx_id_data}
\end{table}

\begin{table}
\footnotesize
\begin{center}
\begin{tabular}{l|l l l}
\toprule
Source  & \multicolumn{3}{r}{Temporally Dependent} \\ 
& No & Maybe & Yes \\ \midrule
NQ-Passage & 73.3 & 15.6 & 11.1 \\
NQ-Table & 54.6 & 3.7 & 41.6 \\
WebQuestions & 59.9 & 10.4 & 29.9 \\
TyDi-QA (en) & 62.3 & 14.1 & 23.5 \\
MS-Marco & 60.0 & 26.4 & 13.6 \\
\bottomrule
\end{tabular}
\end{center}
\caption{Temporal volatility distribution of raw annotation for various datasets (numbers on the training split).}
\label{tab:stage_1_analysis}
\end{table}


\begin{table}
\footnotesize
\begin{center}
\begin{tabular}{lrr}
\toprule
Source & Y/N Data & Y/N/M Data \\ \midrule
\textsc{temp\_abs} / \textsc{temp\_rel} & 0.64 & 0.62 \\
- NQ-Passage & 0.54 & 0.61 \\
- NQ-Table & 0.69 & 0.68 \\
- WebQuestions & 0.46 & 0.49 \\
- TyDi-QA (en) & 0.64 & 0.58 \\
- MS-Marco & 0.43 & 0.45 \\ \midrule
\textsc{geo} & 0.56 & 0.56 \\
\bottomrule
\end{tabular}
\end{center}
\caption{Interannotator Agreement of Identification examples. We compute Fleiss's kappa all examples where the majority of labels (2 out of 3) are either ``Yes'' or ``No'' (first column) and over all examples (second column) ``Yes'', ``No'', or ``Maybe/Unsure''.}
\label{tab:id_agreement_full}
\end{table}

\begin{table*}
\small
\begin{center}
\begin{tabular}{lcc|rrrrr|rrrr}
\toprule
&\multirow{2}{2em}{Query Mod.} & \multirow{2}{2.2em}{Fine-tuned} &  \multicolumn{5}{c|}{\textsc{temp}} & \multicolumn{4}{c}{\textsc{geo}} \\
&  &  & Static & Samp. & Start & Total  & Any  & Comm. & Rare & Total & Any  \\\midrule
\multirow{3}{3em}{Retrieval (DPR)} &&&  77.2 & 70.2 & 67.9 & 70.4 & 80.1 & 53.8 & 45.8 & 50.0 & 85.0 \\
& \checkmark &&                         71.2 & 72.9 & 72.2 & 72.4 & 79.8 & 76.7 & 79.6 & 78.1 & 87.9 \\
& \checkmark & \checkmark &             76.5 & 74.7 & 72.9 & 74.3 & 81.6 & 76.7 & 79.6 & 78.1 & 87.9\\
\bottomrule
\end{tabular} 
\end{center}\vspace{-0.4em}
\caption{Retrieval performance on situated question answering, measured by answer recall within the top 50 retrieved documents. We also report performance for partitions of the test set, was well as recall measured against the union of all answers from all annotated contexts (\textit{Any}).}
\label{tab:retriever_results}
\end{table*}

\begin{table}[]
\footnotesize
\centering
\begin{tabular}{lllrr}
    \toprule
    & Fine- & Knowledge & Stable & Updated \\ 
    & tuned & Source Date & 267     &  209 \\ \midrule
    \multirow{3}{2.75em}{Retrieval (DPR)} &  & Wiki. Dec 2018 & 74.9 & 35.4 \\
    &                                        & Wiki. Feb 2021 & 78.3 & 60.3 \\
    & \checkmark                             & Wiki. Feb 2021 & 79.4 & 66.0 \\
    \bottomrule
\end{tabular}
\caption{Retriever results on the default open-retrieval QA setting where all questions are assumed to be about the present timestamp (Feb 2021), given as answer recall at top 50 on our test set. We split examples based on whether the current answer started to be true \textit{after} the large scale training dataset was collected, \textit{Updated}, or whether questions where answers didn't change since the data collection, \textit{Stable}. Retrieval is run over English Wikipedia dumps from 2018-12-20 and 2021-02-20.}
\label{tab:retriever_corpora_update}
\end{table}

\paragraph{Generating \textsc{temp} $(q,c,a)$ examples}
We sample values of $c_v$ to match the granularity of the annotated dates (year-month-day or year) from our collected answer timelines. For each static question, we uniformly sample a single value of $c_v$ between the original dataset's creation (2018) and the present date (2021-03-01), matching the granularity distribution we find in our training set annotations.




\paragraph{Inter Annotator Agreement}
Inter annotator agreement during identification phase can be found in Table~\ref{tab:id_agreement_full}.
At the validation phase, 70\% of temporal question-context-answer pairs were annotated as correct and 85\% of geographical question-context-answer pairs were annotated as correct, similar to validation phase (76\%) in AmbigQA~\cite{min2020ambigqa}.

\paragraph{Agreement with Natural Questions}
The answer span exact match between our collected answers and the original answers from NQ where the annotated start and end transition dates overlap with NQ's creation (2018) was around 40\%, similar to the average agreement rate on NQ-Open test data is 49.2\% in the original study~\cite{Kwiatkowski2019NaturalQA}. Our manual analysis on randomly sampled 50 errors and find that only 12\% of the errors can be attributed to annotation errors. 70\% of other errors are due to ambiguities in the question resulting in multiple possible answers (e.g. ``Three movies made from Agatha Christie's novels''), or the same answer being given in different forms (e.g. ``Chamberlain, Wilt'' vs. ``Wilt Chamberlain''). The remaining errors are the result of inconsistencies in NQ annotations (18\%).

\section{Data Analysis}\label{sec:appendix_data_analysis}
\paragraph{Naturally occurring query modifications}
In open retrieval QA datasets, questions often ask about the answer in some specific temporal or geographical context. For instance, when people want to ask about the previous answer to a questions, people often add words like "previously" or "last" in their question to specify it. We find that such questions comprise about 4.1\% of NQ-Open~\cite{Kwiatkowski2019NaturalQA,lee-etal-2019-latent}. We also find that 5.1\% of questions specify a specific point in time, as determined by whether there is a year expression in the question. Finally, we estimate the number of questions that specify a geographical context by counting number of stripped questions that were specified as geographically dependent. These questions, which have a phrase specifying the geographical context, comprise about 4.4\% of NQ-Open. Our modifications closely resemble these naturally occurring questions, but can sometimes produce ungrammatical sentences, usually due verb tense agreement.

\section{Retriever Analysis}\label{sec:retriever_analysis}
\paragraph*{Are errors in retrieval based systems due to poor retrieval?}
In Table~\ref{tab:retriever_results}, we report the retrieval performance from the settings in Section~\ref{sec:results}. We measure comparing against the results from Table~\ref{tab:qa_results} and Table~\ref{tab:qa_error_analysis}, we see similar trends in end-to-end performance reflected in our retriever performance.

We also explore whether the retriever model is able to adapt to updated corpora, reporting retrieval performance in Table~\ref{tab:retriever_corpora_update}. Here, we see that while retriever performance also suffers on queries with updated answers, suggesting that retriever systems are perhaps implicitly learning to situate questions within the time period of their large-scale training datasets.

\section{Implementation Details}\label{sec:appendix_implementation}
\paragraph*{Context Dependent Question Identification Baselines}
We use the \texttt{pytorch-transformers} \cite{wolf-etal-2020-transformers} library to implement our classification models. The training batch size is set to 8 and 64 for BERT-base and BERT-large, respectively. We train for 10 epochs using a learning rage of 5e-5 and 500 warmup steps and select the best preforming model measured on dev after each epoch.

\paragraph*{Context Dependent Question Answering Baselines}
We finetune our closed book baselines for 10 epochs with a batch size of 256, using the AdamW optimizer with a learning rate of 1e-5. We select the best performing model evaluated after each training epoch. We keep all other hyperparameters the same from the original implementation.

For our retrieval based baselines, we finetune both the retriever and reader components. The retriever models are trained using in-batch negatives plus one hard-negative passage per question. We use hard-negative and gold passages from the query modified model's predictions before finetuning. We train the retriever model for 8 epochs, using a learning rate of 1e-5, 100 warmup steps, and a batch size of 16. We then finetune our reader models for 16 epochs and a batch size of 16. We select the best model evaluated after each training epoch for both reader and retriever models, and select the best top-$k$ retrieval results to use for in $k \in \{10, 20, 50\}$ evaluated on the development set. All other hyperparameters from the DPR reader and retriever models are kept the same from the original work.

\end{document}